\PassOptionsToPackage{dvipsnames}{xcolor}
\documentclass{article}
\usepackage{iclr2027_conference,times}
\usepackage[T1]{fontenc}

\usepackage{geometry}
\usepackage{amsmath,amssymb}
\usepackage{booktabs}
\usepackage{graphicx}
\usepackage{multirow}
\usepackage{tabularx}
\usepackage{adjustbox}
\usepackage{xcolor}
\usepackage{xspace}
\usepackage{microtype}
\usepackage{marvosym}
\usepackage[format=plain,labelformat=simple,labelsep=period,font=small]{caption}
\usepackage[font=footnotesize,skip=3pt,subrefformat=parens]{subcaption}
\usepackage[shortlabels,inline]{enumitem}
\setlist[itemize]{noitemsep,leftmargin=*,topsep=0em}
\setlist[enumerate]{noitemsep,leftmargin=*,topsep=0em}
\usepackage[most]{tcolorbox}
\usepackage[scaled=0.98]{helvet}

\setcitestyle{numbers,square,sort&compress}

\makeatletter
\DeclareRobustCommand\onedot{\futurelet\@let@token\@onedot}
\def\@onedot{\ifx\@let@token.\else.\null\fi\xspace}
\def\eg{e.g\onedot}

\makeatother

\usepackage{colortbl}
\definecolor{grouplab}{HTML}{6E6E73}
\definecolor{oursrow}{HTML}{F2EFF7}
\definecolor{muted}{HTML}{666666}
\newcommand{\grouphead}[2]{\addlinespace[1.5pt]%
  \multicolumn{#1}{@{}l@{}}{\textcolor{grouplab}{\scshape\footnotesize #2}}\\[1.5pt]}
\newcommand{\ours}{\rowcolor{oursrow}}
\newcommand{\best}[1]{\textbf{#1}}
\newcommand{\cmark}{\ensuremath{\checkmark}}
\newcommand{\xmark}{\ensuremath{\times}}
\newcommand{\pmark}{\ensuremath{\triangle}}

\usepackage{pgfplots}
\pgfplotsset{compat=1.17}
\usetikzlibrary{arrows.meta}

\usepackage{listings}
\usepackage[scaled=0.86]{DejaVuSansMono}
\definecolor{leapviolet}{HTML}{6051C6}
\definecolor{leapgrey}{HTML}{6E6E73}
\definecolor{leapstr}{HTML}{B5562B}
\definecolor{leapmint}{HTML}{7DF2CF}
\newcommand{\leapmark}{\raisebox{-0.14em}{\begin{tikzpicture}[x=0.01em, y=0.01em, line cap=round, line join=round]
  \fill[leapviolet, rounded corners=0.24em] (0,0) rectangle (104,104);
  \fill[white, rounded corners=0.03em] (16,14) rectangle (46,24);
  \draw[white, line width=0.09em] (31,24) -- (33,60) -- (68,74);
  \fill[white] (33,60) circle (7); \fill[leapviolet] (33,60) circle (2.6);
  \draw[white, line width=0.06em] (68,74) -- (68,62) (58,52) -- (58,62) -- (78,62) -- (78,52);
  \draw[leapmint, line width=0.085em] (55,35) -- (64,26) -- (84,46);
\end{tikzpicture}}}
\lstdefinelanguage{Lean4}{
  morekeywords={def,let,do,structure,where,theorem,by,fun,import,open,namespace,end,abbrev,match,with,
    exact,intro,refine,obtain,have,section,variable,if,then,else,pure},
  sensitive=true, morecomment=[l]{--}, morecomment=[n]{/-}{-/}, morestring=[b]"}
\lstdefinestyle{lean}{language=Lean4, basicstyle=\ttfamily\fontsize{6.6}{7.9}\selectfont,
  keywordstyle=\color{leapviolet}\bfseries, commentstyle=\color{leapgrey}\itshape, stringstyle=\color{leapstr},
  columns=fixed, keepspaces=true, showstringspaces=false, upquote=true, extendedchars=true,
  breaklines=true, breakatwhitespace=true, postbreak=\mbox{\hspace{1.5em}},
  literate={←}{{$\leftarrow$}}1 {→}{{$\rightarrow$}}1 {⟨}{{$\langle$}}1 {⟩}{{$\rangle$}}1
    {∀}{{$\forall$}}1 {∈}{{$\in$}}1 {∧}{{$\wedge$}}1 {≤}{{$\le$}}1 {≥}{{$\ge$}}1 {·}{{$\cdot$}}1}
\newtcblisting{leanbox}[2]{listing only, enhanced, breakable=false, colframe=leapviolet, colback=white,
  boxrule=0.6pt, arc=1.6mm, left=5pt, right=5pt, top=1.5pt, bottom=1.5pt, toptitle=2pt, bottomtitle=1.5pt,
  coltitle=leapviolet, colbacktitle=white, fonttitle=\sffamily\bfseries\small,
  titlerule style={leapviolet!30, line width=0.4pt},
  title={#1\hfill{\mdseries\footnotesize\color{leapgrey}#2}}, listing options={style=lean}}

\definecolor{titleboxbg}{HTML}{F3F0FA}
\definecolor{titleboxborder}{HTML}{CFC4E8}
\definecolor{titleink}{HTML}{2E2A36}
\definecolor{linkviolet}{HTML}{6E59B3}
\definecolor{urlmagenta}{HTML}{C93D8E}
\usepackage{hyperref}
\usepackage{url}
\hypersetup{
    colorlinks=true,
    linkcolor=linkviolet,
    citecolor=linkviolet,
    urlcolor=urlmagenta,
    pdftitle={LEAP: Lean as Policy for Verifiable Robot Control},
    pdfauthor={Weichen Fan, Kairui Hu, Haiwen Diao, Siyuan Hu, Ziyi Wang, Ziqi Huang, Ziwei Liu}
}
\usepackage[capitalize]{cleveref}
\crefname{section}{Sec.}{Secs.}
\Crefname{section}{Section}{Sections}
\Crefname{table}{Table}{Tables}
\crefname{table}{Tab.}{Tabs.}

\makeatletter\renewcommand\paragraph{\@startsection{paragraph}{4}{\z@}{0.3ex plus 0.2ex minus .1ex}{-1em}{\normalsize\bf}}\makeatother

\iclrfinalcopy

\begin{document}
\fancyhead{}
\renewcommand{\headrulewidth}{0pt}

\begin{tcolorbox}[
    colback=titleboxbg,
    colframe=titleboxborder,
    boxrule=0.55pt,
    arc=8mm,
    width=\textwidth,
    left=20pt,
    right=20pt,
    top=18pt,
    bottom=16pt,
    before skip=0pt,
    after skip=12pt
]
\color{titleink}

{\color{black}\sffamily\bfseries\fontsize{19}{23}\selectfont
\leapmark\,LEAP:\\ Lean as Policy for Verifiable Robot Control\par}
\vspace{0.80em}

{\color{black}\sffamily\bfseries\fontsize{8.8}{11}\selectfont\raggedright
\mbox{Weichen Fan}\textsuperscript{1},\hspace{0.4em}
\mbox{Kairui Hu}\textsuperscript{1},\hspace{0.4em}
\mbox{Haiwen Diao}\textsuperscript{1},\hspace{0.4em}
\mbox{Siyuan Hu}\textsuperscript{1},\hspace{0.4em}
\mbox{Ziyi Wang}\textsuperscript{1},\hspace{0.4em}
\mbox{Ziqi Huang}\textsuperscript{1},\hspace{0.4em}
\mbox{Ziwei Liu}\textsuperscript{1,\ensuremath{\dagger}}\par}
\vspace{0.45em}

{\fontsize{10.2}{12.4}\selectfont\raggedright
\textsuperscript{1}S-Lab, Nanyang Technological University\par}
\vspace{0.2em}

{\fontsize{8.4}{10.2}\selectfont
\textsuperscript{\ensuremath{\dagger}}Corresponding author.\par}
\vspace{0.70em}

\begingroup
\fontsize{10.5}{13.2}\selectfont
\setlength{\parindent}{0pt}
\setlength{\parskip}{4pt plus 1pt minus 1pt}
\renewenvironment{abstract}{\par\noindent\ignorespaces}{\par}
\begin{abstract}
A manipulation task such as picking up a mug admits many valid actions: the mug can be grasped by its handle or its rim and lifted along countless paths. Yet vision-language-action models, Code-as-Policies and human teleoperation each generate a single action whose validity is found only by running it on the robot. We argue for the opposite order: first define all valid actions, then choose one. We introduce \textbf{LEAP (Lean-as-Policy)}, in which a language-model agent writes the feasible set of a task as a Lean~4 action family over a shared library of physical laws, separating goal, physics and embodiment constraints from user preferences. An objective selects one member of the set, which is checked against the same specification before the robot moves, using exact interval arithmetic and, optionally, the Lean kernel; the check covers only what the family states. Lean thus serves as a unified representation of actions. We instantiate it in LEAP-Harness, which writes feasible actions in Lean from observations and an instruction, then samples verified joint commands. We validate its effectiveness on RoboDojo, RoboCasa365, and LIBERO-Pro.
\end{abstract}

\endgroup

\vspace{0.5em}
{\sffamily\bfseries\fontsize{8}{11}\selectfont
Correspondence:\quad
\url{ziwei.liu@ntu.edu.sg}\par}
\vspace{0.2em}
{\sffamily\bfseries\fontsize{8}{11}\selectfont
Project Page:\quad\url{https://lean-as-policy.github.io/}\par}
\vspace{0.2em}
{\sffamily\bfseries\fontsize{8}{11}\selectfont
Code:\quad\url{https://github.com/WeichenFan/LEAP}\par}

\end{tcolorbox}

\begin{figure}[h]
  \centering
  \includegraphics[width=\textwidth]{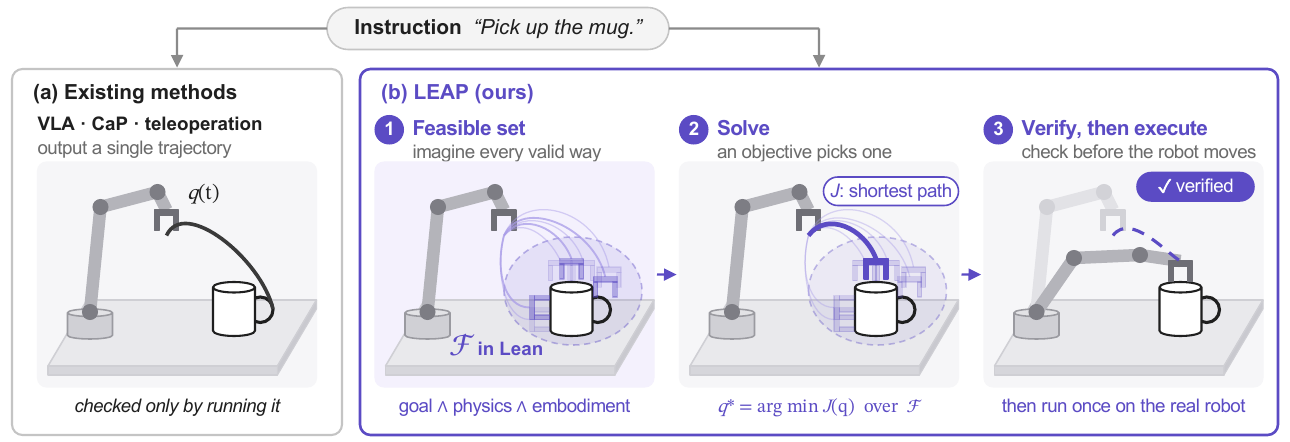}
  \caption{\textbf{LEAP defines all valid actions, picks one, and verifies it before execution.}
  (a)~VLA models, Code-as-Policies (CaP) and human teleoperators each output a single trajectory $q(t)$,
  whose validity is found only by running it.
  (b)~LEAP \textcircled{\scriptsize 1} states the feasible set $\mathcal{F}$ in Lean as goal, physics and embodiment
  constraints, which admits every valid way to perform the task (shaded region and translucent grippers: rim, handle or side grasps, each with a continuum of poses); \textcircled{\scriptsize 2} lets an objective $J$ select
  $q^\ast = \arg\min_{q \in \mathcal{F}} J(q)$; and \textcircled{\scriptsize 3} checks $q^\ast$ against all constraints of $\mathcal{F}$
  before running it once.}
  \label{fig:teaser}
\end{figure}

\section{Introduction}
\label{sec:intro}

\looseness=-1\relax Ask a robot to pick up a mug, and there is no single correct action. The mug can be grasped by its handle or by its rim, at many points along either, and lifted along countless paths. They differ only in properties such as path length or clearance, which reflect a user's preference, not the task. Physically, a manipulation task is a set of trajectories: those that achieve the goal while respecting contact physics and the robot's limits. Yet current approaches output one member of that set without stating the set or checking the member against it.

\looseness=-1\relax Vision-language-action (VLA) models~\cite{brohan2023rt2,kim2024openvla,black2024pi0,intelligence2025pi05} map observations directly to trajectories. They are the System~1~\cite{kahneman2011thinking} of manipulation: fluent on their training distribution, but they learn which trajectories were demonstrated rather than why those trajectories succeed, and they degrade when the scene, the objects, or the embodiment shift~\cite{zhou2025liberopro}. Language-model planners deliberate at a higher level, composing semantic commands such as \texttt{move\_to} and \texttt{pick}~\cite{huang2022zeroshot,ahn2022saycan,vemprala2024chatgpt}; Code-as-Policies (CaP)~\cite{liang2023cap} goes further, mapping instructions to code that reacts to perceptual inputs, parameterizes control-primitive APIs, and is executed directly on the robot. These methods each generate one action (Fig.~\ref{fig:teaser}a): a VLA decodes one trajectory from the current observation, a CaP program commits to one sequence of primitive calls with fixed parameters, and a human teleoperator demonstrates one path. Whether that action works is found out only by executing it. None of these methods represents the set of acceptable actions, so an action cannot be checked before the robot moves; in a new scene, the robot itself is the only test. And when two actions both succeed, nothing says which is better, because the subjective preference that would decide is never stated.

\looseness=-1\relax We argue for the opposite order (Fig.~\ref{fig:teaser}b): first define all valid actions as a set, then sample one from it. This puts the feasible set at the top of the action hierarchy, feasible set $\supseteq$ semantic command $\supseteq$ trajectory: a command such as \texttt{grasp(mug, handle)} selects the subset reachable by one strategy, and a trajectory is a single member. The set is also continuous: where a generated action commits to a single grasp point, the set states that any grasp point within a 1--2\,cm interval of the handle is valid. Stating the set separates which actions are acceptable, a question that can be checked, from which acceptable action to take, an optimization the user controls. Formal mathematics shows why this split matters: a language model often proposes wrong proofs, but Lean checks each one mechanically, so every proof it accepts is correct~\cite{hubert2025alphaproof,ren2025deepseekproverv2}. Robot manipulation has so far lacked such a mechanical checker for its actions.

We introduce \textbf{LEAP} (Lean-as-Policy), which writes the feasible set of a manipulation task as an action family in Lean~4~\cite{demoura2021lean4}. Given the instruction, what perception sees, and a model of the robot, a language-model agent writes a short Lean program that describes the valid actions instead of producing one. The program reuses a shared library of physical laws (kinematics, geometry, contact, friction, and dynamics) and states three kinds of constraints: what the instruction asks for (\eg, the mug ends up lifted), what physics allows (\eg, the grasp does not slip), and what the robot can execute (\eg, its joint limits). Quantities that perception cannot pin down exactly, such as the position of the mug, enter as intervals rather than single numbers. Preferences such as speed or clearance stay out of the constraints and form a separate objective. The agent therefore never writes a joint target: it states what a correct action is, and a solver finds one that satisfies it.

\looseness=-1\relax This gives LEAP two properties. First, LEAP acts like a simulator for the task: any candidate action, from any source, can be verified before the robot moves. LEAP checks the candidate against every constraint for all values that perception allows; if the check passes, the action is certified, and Lean can also prove this formally. Second, because LEAP defines the whole family of valid actions, a trajectory can be drawn from it flexibly: given an objective, it can be found by a numerical solver, by probabilistic sampling, or even by a learned model.

\looseness=-1\relax We implement LEAP as \textbf{LEAP-Harness}, a five-module pipeline. Perception segments the scene with perception tools
 like SAM~3~\cite{carion2026sam3} and recovers its 3D geometry with Depth Anything~3~\cite{lin2026da3} or Stereo depth, recording how uncertain each measured quantity is. 
The Lean Formalizer, a language-model agent, writes the action family in Lean. The Solver searches the family for the trajectory that best meets the objective. The Verifier checks that trajectory against every constraint, optionally with a Lean proof. The Executor then runs the verified trajectory on the robot once. If no valid trajectory is found, the agent works out which constraint failed, gathers more evidence, and revises the family instead of retrying blindly. We evaluate LEAP-Harness mainly on the simulation track of RoboDojo~\cite{chen2026robodojo}, whose 42 tasks span generalization, precision, long-horizon, memory, and open-vocabulary manipulation, and further on RoboCasa365~\cite{nasiriany2026robocasa365} and LIBERO-PRO~\cite{zhou2025liberopro}, and on three real tabletop tasks with a robot arm in our lab.

Our main contributions are:
\begin{itemize}[leftmargin=*,topsep=2pt,itemsep=1pt,parsep=0pt]
  \item \textbf{A Lean-defined action family} that states the feasible set of a manipulation task over a library of physical laws, separating goal, physics, and embodiment constraints from preferences, so that any candidate action, from any source, can be verified with an explicit scope.
  \item \textbf{A flexible sampler} that searches the family under any objective, estimated success or a user reward, and distinguishes plans certified for every value perception allows from plans that are only estimated.
  \item \textbf{LEAP-Harness}, an agent harness that performs robot manipulation in the Lean-as-Policy manner, which we evaluate on three simulated benchmarks and a real robot.
\end{itemize}

\section{Related Work}
\label{sec:related}

Robot policies built on foundation models differ in what they output and in when that output is demonstrated in \cref{tab:paradigms}. A VLA outputs an action chunk from its weights, an agent harness outputs one tool call or code cell at a time, and a code policy outputs one program. In all three, an action is judged only by running it on the robot, so an invalid action is discovered only after it fails.

\begin{table}[t]
  \centering
  \footnotesize
  \setlength{\tabcolsep}{1.8pt}
  \begin{tabularx}{\textwidth}{@{}l*{4}{>{\centering\arraybackslash}X}@{}}
    \toprule
    & VLA & Agent & Code & \textbf{LEAP} \\
    \midrule
    Output & Trajectory & Trajectory & Trajectory & Action Set \\
    Valid actions stated & \xmark & \xmark & \xmark & \cmark \\
    Checked before execution & \xmark & \xmark & \pmark & \cmark \\
    Physics with units & \xmark & \xmark & \xmark & \cmark \\
    Uncertainty as intervals & \xmark & \xmark & \xmark & \cmark \\
    Preference separate & \xmark & \xmark & \xmark & \cmark \\
    Model calls & per step & per call & 0 to per cell & per family \\
    \bottomrule
  \end{tabularx}
  \caption{\textbf{Four ways to build a robot policy with foundation models.} \cmark\ yes, \pmark\ in part, \xmark\ no. VLA: vision-language-action model. Agent: agent harness. Code: code policy, whose programs may be tested on fixtures before deployment. Checked before execution: an action is tested against a stated specification before the robot runs it, not only by running it.}
    \label{tab:paradigms}
\end{table}

\paragraph{Vision-language-action models.}
\looseness=-1\relax Diffusion Policy~\cite{chi2023diffusion}, RT-1~\cite{brohan2023rt1}, RT-2~\cite{brohan2023rt2}, Octo~\cite{ghosh2024octo}, OpenVLA~\cite{kim2024openvla}, $\pi_0$, and $\pi_{0.5}$~\cite{black2024pi0,intelligence2025pi05} learn visuomotor policies from demonstrations. Which actions are valid is never written down; it is implicit in the data and the weights. LIBERO-PRO shows that high benchmark success can reflect memorized configurations~\cite{zhou2025liberopro}. LEAP instead defines the valid action set in Lean, so a VLA's action can be checked against it, and a model can be trained within LEAP, \eg, to write action families or to propose actions for the verifier.

\paragraph{Language-model agents for robot control.}
SayCan grounds language-model plans in learned affordances~\cite{ahn2022saycan}, Inner Monologue feeds execution feedback back to the planner~\cite{huang2022inner}, and other work composes skills from language~\cite{huang2022zeroshot,vemprala2024chatgpt}. Recent harnesses let an agent call analytic or learned primitives during the episode: AMP1 runs one harness across embodiments~\cite{chen2026amp1}, and GPT-6 Astra acts as a direct end-effector policy on RoboDojo~\cite{zhang2026astra}. The model decides each action, and the action is tested by running it. In LEAP, the agent writes only constraints; a model-free solver and verifier produce and check the action before it runs.

\paragraph{Code as policy.}
Code as Policies and ProgPrompt prompt a language model to write programs over perception and control APIs~\cite{liang2023cap,singh2023progprompt}, and VoxPoser writes code that composes spatial value maps~\cite{huang2023voxposer}. CaP-X benchmarks coding agents across primitive levels and interaction modes~\cite{fu2026capx}. RHO and HarnessPAI evolve programs across rollouts~\cite{elmaaroufi2026rho,darwin2026harnesspai}. PyRUA-Lean has a model write code cell by cell as each episode runs~\cite{si2026pyrua}. Code-Only-as-Policy (COAP) has coding agents write a shared library offline that measures the robot and scene state from images and runs on RoboDojo with no model in the loop~\cite{hu2026etm}. Each program encodes one procedure, checked after a step has run or on fixtures offline; a LEAP family states the valid motions, and each candidate, from any source, is checked against it before the robot moves.

\paragraph{Constraints, planning, and formal methods.}
Task and motion planning describes manipulation as constraints~\cite{garrett2021tamp}, solved by sampling~\cite{garrett2020pddlstream} or optimization~\cite{toussaint2015lgp}. Text2Motion checks the geometric feasibility of skill sequences~\cite{lin2023text2motion}, LLM+P and AutoTAMP translate language into PDDL or temporal logic~\cite{liu2023llmp,chen2024autotamp}, ReKep has a vision-language model write keypoint constraints as Python costs~\cite{huang2024rekep}, and PRoC3S samples the open parameters of a language-model program to satisfy its constraints~\cite{curtis2024proc3s}. In these methods, the constraints are a solver's internal problem, and its output, one plan or trajectory, is what is kept and executed. In LEAP, the output is the set of valid actions itself: a typed Lean family over physical laws with checked units and interval beliefs, against which every candidate is verified over all scenes the beliefs allow, and which persists to judge actions from any source (\cref{sec:exp:robot,sec:exp:data}). Formal specifications have long been used to synthesize robot controllers from temporal logic~\cite{kress2009temporal}, and language models paired with Lean~\cite{demoura2021lean4} now produce machine-checked proofs~\cite{yang2023leandojo,hubert2025alphaproof,ren2025deepseekproverv2}. LEAP applies the same split between a generator and a checker to physical actions.

\section{LEAP: Lean as Policy}
\label{sec:method}

A policy usually maps an observation to one action; LEAP maps it to the set of valid actions, stated in Lean, and takes the action from it. Each decision has five steps: observe the scene as beliefs (\cref{sec:method:observe}), formulate the task in Lean (\cref{sec:method:lean}), solve for a candidate (\cref{sec:method:solve}), verify it against the specification (\cref{sec:method:verify}), and execute it (\cref{sec:method:execute}). Only the formulation needs judgment: a prompted agent (\cref{sec:harness}), a trained model, or a person can write it; the other steps use no model. Our example is ``Pick up the mug'' (\cref{fig:teaser}).

\begin{figure*}[t]
  \centering
  \includegraphics[width=\textwidth]{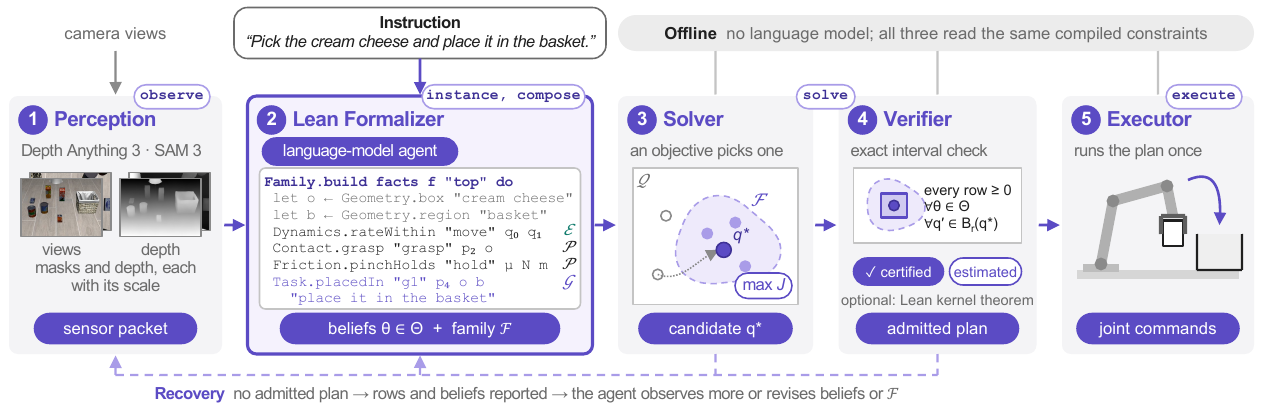}
  \caption{\looseness=-1\relax \textbf{LEAP-Harness.} Five modules carry out the steps of \cref{sec:method}; tabs name the agent tools that invoke them. The solver, verifier, and executor run offline on the same compiled constraints; without an admitted plan, the agent observes more or revises (dashed). Shown on a LIBERO-PRO episode, with Depth Anything~3 (Small) depth. The Lean excerpt is abridged; $\mathcal{E}$, $\mathcal{P}$, $\mathcal{G}$: the embodiment, physics, and goal rows.}
  \label{fig:harness}
\end{figure*}

\subsection{Observe}
\label{sec:method:observe}

\looseness=-1\relax Perception turns images and robot state into beliefs. A belief describes one physical quantity, such as the position of the mug or the friction coefficient of the gripper pads, as an interval with the evidence it came from and optionally a distribution, such as the cup's mass in $[0.32, 0.40]$\,kg (\cref{fig:cup}b). A quantity that nothing observed stays unknown, and a prior remains a prior. A scene $\theta$ assigns one value to every quantity; $\Theta$ is the set of scenes the beliefs allow, with every quantity anywhere in its interval. Observations enter the specification only through these beliefs, never as raw pixels.

\subsection{Formulate in Lean}
\label{sec:method:lean}

\begin{figure}[t]
  \centering
  \begin{tikzpicture}[x=1cm, y=1cm, >=Stealth, font=\sffamily\footnotesize]
  \node[font=\sffamily\bfseries\footnotesize, anchor=south west] at (-3.4,3.05) {(a) Beliefs};
  \draw[leapgrey!70, fill=leapgrey!6, rounded corners=2pt] (-3.3,0.2) rectangle (-1.0,2.85);
  \node[leapgrey, anchor=north east] at (-1.05,2.82) {$\Theta$};
  \fill[leapgrey] (-2.75,2.15) circle (2.2pt) node[left=1pt] {$\theta_1$};
  \fill[leapstr] (-1.65,0.8) circle (2.2pt) node[right=1pt] {$\theta_2$};
  \fill[teal!80!black] (-1.95,2.2) circle (2.2pt) node[below left=-1pt] {$\theta_3$};
  \node[leapgrey, align=center, font=\sffamily\scriptsize] at (-2.15,-0.08) {scenes perception allows};
  \draw[->, leapgrey, thick] (-0.85,1.7) -- node[above, font=\sffamily\scriptsize] {each $\theta$} node[below, font=\sffamily\scriptsize] {gives $\mathcal{F}(\theta)$} (0.15,1.7);
  \node[font=\sffamily\bfseries\footnotesize, anchor=south west] at (0.2,3.05) {(b) Action space $\mathcal{Q}$};
  \draw[leapgrey!70, rounded corners=2pt] (0.3,-0.35) rectangle (7.3,3.0);
  \def\Ea{(3.25,1.6) ellipse [x radius=2.15, y radius=1.15, rotate=8]}
  \def\Eb{(4.35,1.45) ellipse [x radius=2.0, y radius=1.05, rotate=-12]}
  \def\Ec{(3.95,1.95) ellipse [x radius=1.75, y radius=0.85]}
  \begin{scope} \clip \Ea; \clip \Eb; \fill[leapviolet!30] \Ec; \end{scope}
  \draw[leapgrey, thick] \Ea; \draw[leapstr, thick] \Eb; \draw[teal!80!black, thick] \Ec;
  \node[leapgrey] at (1.15,2.35) {$\mathcal{F}(\theta_1)$};
  \node[leapstr] at (6.45,0.65) {$\mathcal{F}(\theta_2)$};
  \node[teal!80!black] at (5.35,2.72) {$\mathcal{F}(\theta_3)$};
  \node[leapviolet, font=\sffamily\bfseries\footnotesize] at (3.95,2.08) {$\mathcal{F}_{\mathrm{rob}}$};
  \draw[leapviolet, densely dashed] (3.75,1.4) circle (0.26);
  \fill[leapviolet] (3.75,1.4) circle (1.6pt);
  \draw[leapviolet] (3.93,1.3) -- (4.6,0.02);
  \node[leapviolet, anchor=east, font=\sffamily\scriptsize, fill=white, inner sep=1.5pt] at (7.22,-0.08) {$q^\ast$: $B_r(q^\ast) \subseteq \mathcal{F}_{\mathrm{rob}}$, certified};
  \draw[leapgrey, densely dashed] (1.65,1.3) circle (0.26);
  \fill[leapgrey] (1.65,1.3) circle (1.6pt);
  \draw[leapgrey] (1.6,1.04) -- (1.3,-0.05);
  \node[leapgrey, anchor=west, font=\sffamily\scriptsize, fill=white, inner sep=1.5pt] at (0.4,-0.08) {$q' \notin \mathcal{F}(\theta_2)$};
\end{tikzpicture}
  \caption{\textbf{The robust feasible set.} (a)~Perception leaves a set $\Theta$ of scenes consistent with the beliefs, of which three are shown. (b)~Each scene $\theta \in \Theta$ induces a feasible set $\mathcal{F}(\theta)$ (\cref{eq:feasible}), and $\mathcal{F}_{\mathrm{rob}}$ is their intersection (\cref{eq:robust}). A candidate $q$ is certified if its execution set satisfies $B_r(q) \subseteq \mathcal{F}_{\mathrm{rob}}$, as for $q^\ast$. The candidate $q'$ satisfies every row in scene $\theta_1$ but not in $\theta_2$; hence $q' \notin \mathcal{F}_{\mathrm{rob}}$, and the verifier can at most report a chance estimate for it.}
  \label{fig:robust}
\end{figure}
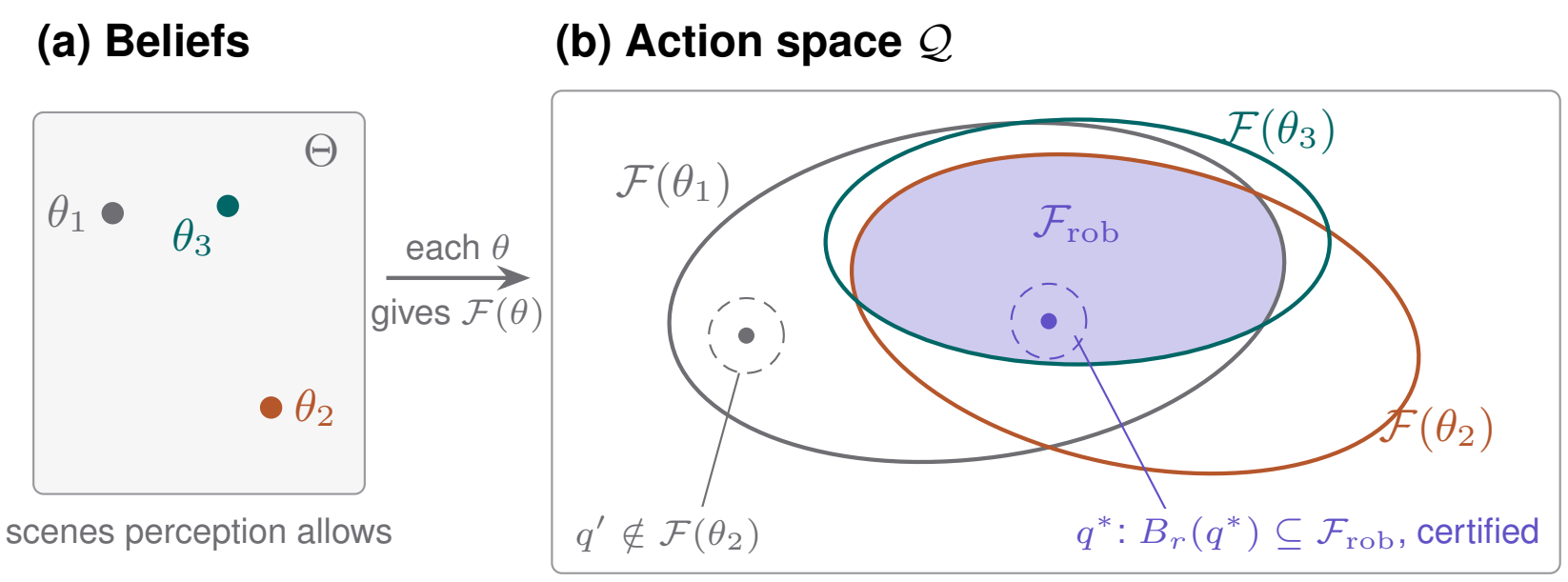

\paragraph{Actions and the feasible set.}
An action $q \in \mathcal{Q}$ is what the robot executes from its current state $q_0$: a trajectory $q(t)$, joint configurations with gripper events, or another form a family declares, so LEAP assumes neither discrete nor continuous actions. A task constrains actions through rows $g_i(q, \theta) \geq 0$: goal rows $\mathcal{G}$ encode the instruction, physics rows $\mathcal{P}$ what contact and dynamics require, and embodiment rows $\mathcal{E}$ what the robot can execute. Execution is inexact: when $q$ is commanded, the robot performs some action in an execution set $B_r(q) \subseteq \mathcal{Q}$, where $r$ is the tolerance of the executor; the simplest instance is the ball of radius $r$ about $q$. We require every $q' \in B_r(q)$ to satisfy all rows. For one scene $\theta$, the feasible set is
\begin{equation}
  \mathcal{F}(\theta) = \bigl\{\, q \in \mathcal{Q} :\; g_i(q', \theta) \geq 0 \;\; \forall\, i \in \mathcal{G} \cup \mathcal{P} \cup \mathcal{E},\;\; \forall\, q' \in B_r(q) \,\bigr\},
  \label{eq:feasible}
\end{equation}
and the actions that are valid for every scene consistent with the beliefs form the robust feasible set (\cref{fig:robust})
\begin{equation}
  \mathcal{F}_{\mathrm{rob}} = \textstyle\bigcap_{\theta \in \Theta} \mathcal{F}(\theta).
  \label{eq:robust}
\end{equation}
For the mug, $\mathcal{F}$ contains grasps at the rim, the handle, and the side; a command such as \texttt{grasp(mug, handle)} selects one strategy, and a trajectory is a single member, which gives the hierarchy $\mathcal{F} \supseteq c \supseteq q(t)$ of \cref{fig:teaser}.

\begin{figure*}[t]
  \centering
  \includegraphics[width=\textwidth]{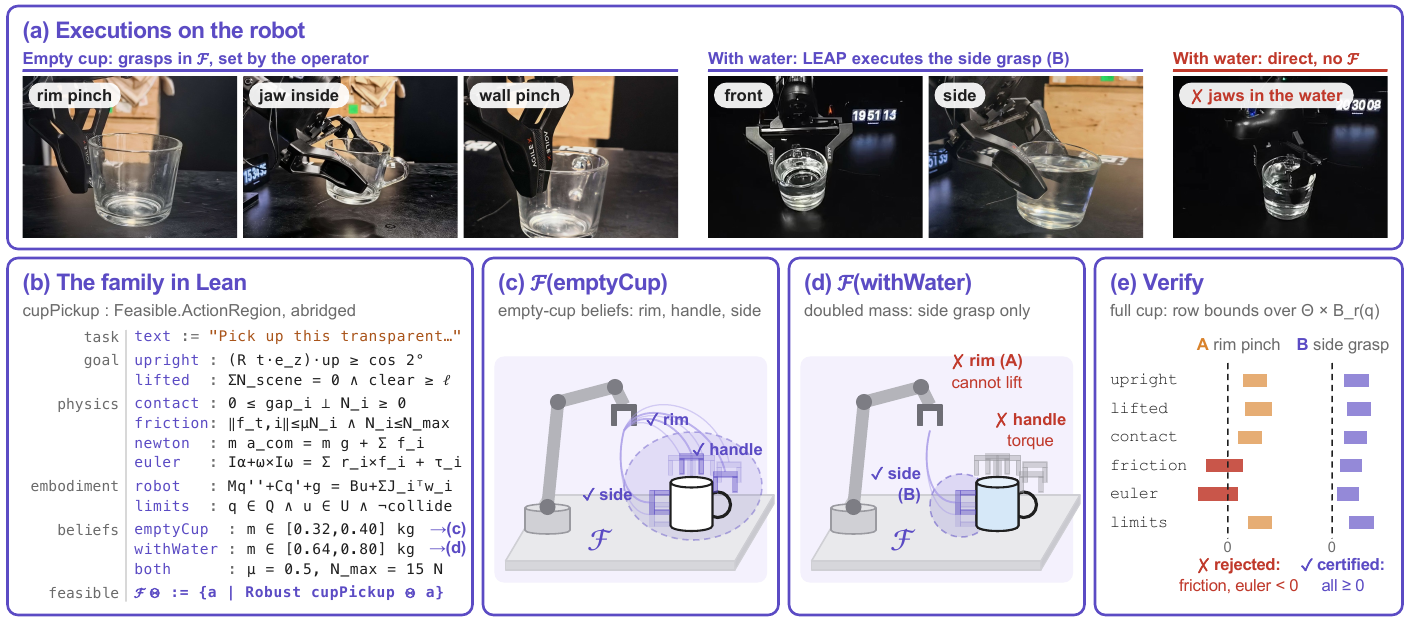}
  \caption{\textbf{Picking up a glass cup, empty and with water.} (a) On the robot: three operator-set grasps of the empty cup in $\mathcal{F}$, LEAP's side grasp of the full cup, and the model acting directly. (b) The family, abridged, over one \texttt{Feasible.ActionRegion}; the two scenes differ only in the mass belief. (c) Under the empty-cup beliefs, $\mathcal{F}$ holds rim, handle, and side grasps. (d) Water doubles the mass prior; only the side grasp (B) remains. (e) The verifier bounds every row over $\Theta$ and the execution sets: A has negative friction and Euler bounds and is rejected; B is certified. Bar values are illustrative; grasp regions follow from the stated rows and a pad-contact model of the PiPER gripper.}
  \label{fig:cup}
\end{figure*}

\paragraph{Action family.}
$\mathcal{F}$ is written as an action family: a short Lean~4 module over a shared library of physical laws. A law is a Lean function that maps expressions and beliefs to rows, together with the premises under which those rows hold; the library covers kinematics, geometry, contact, friction, dynamics, and articulated mechanisms. Laws contain no numbers: masses, friction coefficients, and tolerances enter as beliefs or as literals with units, and a row that mixes dimensions is rejected. The family reads beliefs by name, declares the action's parameters as free variables, calls laws, and adds goal rows that quote the instruction, so it states what makes an action valid without choosing one. A rim grasp of the mug, for example, bounds joint rates and static moments ($\mathcal{E}$), requires the open jaws to straddle the rim and the pinch to hold the load, $2\mu N \geq m g$ ($\mathcal{P}$), and the mug to end lifted ($\mathcal{G}$); handle and side grasps call other contact laws under the same goal, and $\mathcal{F}$ is their union. In Lean, a family is a \texttt{Feasible.ActionRegion}, and \cref{eq:robust} is its \texttt{Robust} predicate (\cref{fig:cup}b).

\paragraph{Human preferences.}
\looseness=-1\relax Hard rows are limited to goals, the equations of the called laws, and the limits of the robot. Preferences enter only an objective $J$ that ranks members of $\mathcal{F}$, by default the estimated probability that $q \in \mathcal{F}(\theta)$ for $\theta$ drawn from the beliefs, or a user's reward; $\mathcal{F}$ thus stays as large as the task allows.

\paragraph{Compiled constraints.}
Lean type-checks the family and compiles its rows into compiled constraints, a program that, given an action $q$ and a scene $\theta$, computes every $g_i(q, \theta)$ exactly, in integer arithmetic without rounding. The solver searches with them, the verifier bounds them, and the executor sends only commands bound to them, so the solved, verified, and executed actions are always one and the same.

\subsection{Solve}
\label{sec:method:solve}

The solver seeks $q^\ast = \arg\max_{q \in \mathcal{F}} J(q)$ from the compiled constraints alone, with no task names or motion templates. It first bounds every hard row by interval arithmetic over all of $\mathcal{Q}$ and $\Theta$; if some row is negative everywhere, the family is certified empty, and the conflicting rows and beliefs are reported. Otherwise it runs local search from $q_0$, its perturbations, and random points of $\mathcal{Q}$, first reducing the violation of the hard rows at the center of $\Theta$ and then raising the smallest row margin (and the reward, if given); the verifier admits candidates, and $J$ selects among them. The search only proposes, and because it is local, an empty result does not prove that $\mathcal{F}$ is empty; a learned sampler or another policy could also propose candidates.

\subsection{Verify}
\label{sec:method:verify}

For a candidate $q$, the verifier computes a lower bound $\underline{g}_i\bigl(B_r(q), \Theta\bigr)$ of each hard row over $B_r(q)$ and all scenes in $\Theta$, using exact interval arithmetic~\cite{moore2009interval} on the integer program; interval bounds are conservative, so a negative bound means the candidate is not certified, not that it fails. If $\underline{g}_i \geq 0$ for every row, then $q \in \mathcal{F}_{\mathrm{rob}}$ (\cref{eq:robust}), and the plan is robustly certified (\cref{fig:cup}e). When the beliefs are too wide for a certificate, the verifier reports on how many of $N$ belief samples every row holds over $B_r(q)$, with a Wilson interval; such a plan is chance-estimated. On request, Lean evaluates the family on the candidate, and its kernel checks a theorem, proved from the checker's soundness using only standard axioms, that every hard row holds over both sets.

\looseness=-1\relax A certificate is relative to the family. It covers the stated rows and their premises, but not whether the beliefs contain the true scene, the robot tracks its commands, the rows hold between checked points, or contact follows the chosen laws. Each plan records the claims it carries, so a failure later on the robot can be traced either to the specification or to the world that it describes.

\subsection{Execute}
\label{sec:method:execute}

An admitted plan, certified or estimated above a threshold, is executed once as joint commands, and its observations update the beliefs for the next step; if none is admitted, the author observes more or revises.

\section{LEAP-Harness}
\label{sec:harness}

\looseness=-1\relax LEAP-Harness instantiates LEAP with a language-model agent as the author of the specification. Its five modules carry out the steps of \cref{sec:method} (\cref{fig:harness}). The agent drives them through the tools \texttt{observe}, \texttt{instance}, \texttt{compose}, \texttt{solve}, and \texttt{execute}, and uses \texttt{read} and \texttt{finish} to inspect results and stop. No tool accepts a joint target, and \texttt{execute} runs only plans the verifier admitted, so every robot motion passes through Lean.

\paragraph{Perception.}
For each camera view, perception runs a depth model, Depth Anything~3~\cite{lin2026da3} by default, and SAM~3~\cite{carion2026sam3}, and stores masks and depth in an immutable sensor packet recording whether each depth map is aligned to known extrinsics, metric by the model alone, or relative. Perception does not interpret the scene: through \texttt{observe}, the agent queries the packet, \eg for plane fits, agreement across views, or the circle of a rim, and decides which detection is which object and how wide each belief is.

\paragraph{Lean Formalizer.}
\looseness=-1\relax This is the only module that uses a language model. Through \texttt{instance}, the agent states the beliefs $\theta \in \Theta$; through \texttt{compose}, it writes the action family $\mathcal{F}$. The host renders the beliefs and the robot model (kinematic chain, joint domains, stop radii, and tool geometry) into a Lean module of facts, against which Lean elaborates the family. The host adds no rows, so the specification contains exactly what the agent and the called laws state. Each family is compiled twice: at a fast fidelity, and at a slow one that adds law detail. 

\paragraph{Solver, verifier, and executor.}
\looseness=-1\relax All three run offline on the same compiled constraints. A call to \texttt{solve} runs the solver and the verifier (\cref{sec:method:solve,sec:method:verify}) under an objective and a fixed budget of starts, iterations, belief samples, and wall time, and returns a candidate and, if admitted, a plan. Each episode fixes an admission policy, certified plans only or estimates above a threshold, and a verification level: the exact integer check, agreement with Lean's own evaluation, or a kernel theorem. A plan binds the family, facts, compiled constraints, candidate, and joint commands, so \texttt{execute} sends exactly the commands that were verified. Otherwise, \texttt{solve} reports the responsible rows and beliefs, and the agent observes more, as its observation policy permits, or revises its beliefs or family but not the goal rows.

\paragraph{Implementation.}
The law library and the checker are written in Lean~4 (v4.19.0) without external mathematical libraries. Benchmarks connect through a common world interface, with adapters for RoboDojo~\cite{chen2026robodojo}, RoboCasa365~\cite{nasiriany2026robocasa365}, and LIBERO-PRO~\cite{zhou2025liberopro}. An observation policy fixes whether the agent observes and executes once or repeatedly in an episode. The agent is Claude Opus 5.5.

\section{Experiments}
\label{sec:experiments}

\looseness=-1\relax We test LEAP on a real robot (\cref{sec:exp:real,sec:exp:robot}) and on LIBERO-PRO (\cref{sec:exp:libero}), convert robot data into Lean targets (\cref{sec:exp:data}), and report benchmarks (\cref{sec:exp:bench}).

\begin{figure}[t]
  \centering
  \begin{minipage}[t]{0.555\textwidth}
    \vspace{0pt}\centering
    \includegraphics[width=\linewidth]{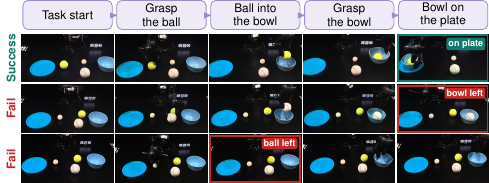}
    \caption{\textbf{Trajectories to judge before execution.} Columns show the end of each step. Before a recorded trajectory runs, LEAP and GPT-6 Astra predict whether it succeeds (\cref{tab:real}). Tennis ball: success; baseball: bowl not lifted; golf ball: grasp missed.}
      \label{fig:real}
  \end{minipage}\hfill
  \begin{minipage}[t]{0.415\textwidth}
    \vspace{0pt}\centering
    \captionsetup{type=table}
    \footnotesize\setlength{\tabcolsep}{3.2pt}
    \begin{tabular}{@{}lcccc@{}}
      \toprule
      & \multicolumn{2}{c}{Success} & \multicolumn{2}{c}{Judge accuracy} \\
      \cmidrule(lr){2-3} \cmidrule(l){4-5}
      Task & Astra & LEAP & Astra & LEAP \\
      \midrule
      Tennis ball & 3/5 & 5/5 & 1/10 & 10/10 \\
      Baseball & 0/5 & 4/5 & 0/10 & 10/10 \\
      Golf ball & 2/5 & 4/5 & 1/10 & 10/10 \\
      \bottomrule
    \end{tabular}
    \vspace{2pt}
    \caption{\textbf{Real-robot results.} Success: completed rollouts, of five per task. Judge: recorded trajectories judged correctly before they run, of ten per task, five successful. Astra: GPT-6 Astra.}
      \label{tab:real}

  \end{minipage}
\end{figure}

\begin{figure*}[t]
  \centering
  \includegraphics[width=\textwidth]{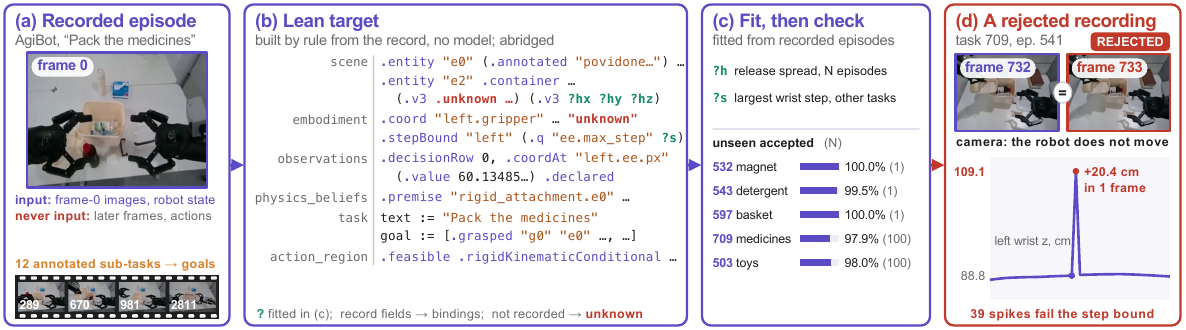}
  \caption{\textbf{From recorded episodes to Lean targets.} (a) An AgiBot episode of task 709: frame-0 observations form the input; the 12 annotated sub-tasks (film strip) become goals. (b) The target, built by rule without a model (abridged): record fields become Lean bindings, unfinished sub-tasks become goals, unrecorded values stay \texttt{unknown}, and two values are left to fit, a container half-extent \texttt{?h} and the step bound \texttt{?s}. (c) \texttt{?h} is the 99\% bound of release offsets from the first release in $N$ calibration episodes, and \texttt{?s} the largest wrist step in other tasks; bars: unseen episodes accepted (mean of 20 draws). (d) A rejected episode (task 709, episode 541): between frames 732 and 733 the camera is unchanged, but the recorded left wrist rises by 20.4\,cm; 39 such spikes fail the step bound, yet the task is completed.}
  \label{fig:data}
\end{figure*}

\subsection{A worked example: picking up a cup}
\label{sec:exp:real}

LEAP-Harness, with GPT-6 Astra~\cite{zhang2026astra} as the agent, controls an AgileX PiPER arm to ``Pick up this transparent cup while keeping its rim level and its body upright'' (\cref{fig:cup}). The family (\cref{fig:cup}b) states contact, friction ($\mu = 0.5$, $N \le 15$\,N), and the Newton--Euler equations of the cup and the arm, keeps the cup within $2^\circ$ of upright, and names no grasp type. For the empty cup, $\mathcal{F}$ contains rim, handle, and side grasps (\cref{fig:cup}a,c), and the objective chooses among them; the physics rows already exclude plausible-looking actions, such as a handle grasp with horizontal jaws, whose pads cannot resist the torque of holding the cup level.

\paragraph{Adaptation through beliefs.}
\looseness=-1\relax Water doubles the mass prior; only the mass belief changes. A rim pinch would now need a 26\,mm instead of a 13\,mm deep grip, and a handle pinch a 43\,mm grip on a 23\,mm bar, leaving only the side grasp; LEAP lifts the full cup (\cref{fig:cup}a,d), and the direct model puts its jaws into the water.

\paragraph{Rejection with a reason.}
Before the robot moves, the verifier rejects a rim pinch, whose friction and Euler rows have negative lower bounds, and certifies a side grasp (\cref{fig:cup}e); the negative bounds name the rows that block certification, which a direct policy discovers only by failing on the robot. The example is a qualitative illustration, not an end-to-end verified execution: the operator set the empty-cup grasps, and for the full cup the 6\,cm lift and the jaw apertures after a 3\,cm Lean check; neither was verified.

\subsection{Capability on a real robot}
\label{sec:exp:robot}

\looseness=-1\relax Three tasks share a tabletop with a plate, a bowl, and three balls: put the tennis ball, the baseball, or the golf ball into the bowl, then the bowl on the plate, so that the second step moves the container of the first (\cref{fig:real}). We compare LEAP-Harness with GPT-6 Astra~\cite{zhang2026astra}, which outputs one end-effector target per model call, based on embodiment information and all the sensors' input. Of 15 rollouts, LEAP completes 13 and GPT-6 Astra 5, none with the baseball (\cref{tab:real}). In the trajectories of \cref{fig:real}, GPT-6 Astra never lifts the bowl with the baseball, and it misses the golf ball but still goes on to the bowl.

\paragraph{Judging recorded trajectories.}
A feasible set can also judge an action that another policy produced. For each task, we record ten trajectories by human teleoperation, five of them successful; before each runs, LEAP checks its joint states against the task's Lean family (\cref{sec:exp:data}) and GPT-6 Astra judges based on the same input. LEAP is right on all 30, GPT-6 Astra on 2, showing that the check transfers to real trajectories (\cref{tab:real}).

\begin{figure}[t]
  \centering
  {\footnotesize\sffamily
  \begin{tabularx}{\textwidth}{@{}*{5}{>{\centering\arraybackslash}X}@{}}
    \multicolumn{5}{@{}l}{\textbf{(a) Lid, card box, container}} \\[1pt]
    \multicolumn{5}{@{}c@{}}{\includegraphics[width=\textwidth]{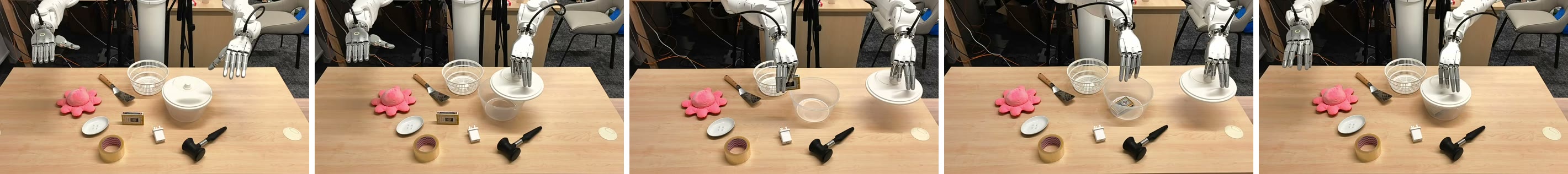}} \\
    Start & Lid lifted and held & Box picked up & Box inside & Lid put back \\[5pt]
    \multicolumn{5}{@{}l}{\textbf{(b) Basket and bowl}} \\[1pt]
    \multicolumn{5}{@{}c@{}}{\includegraphics[width=\textwidth]{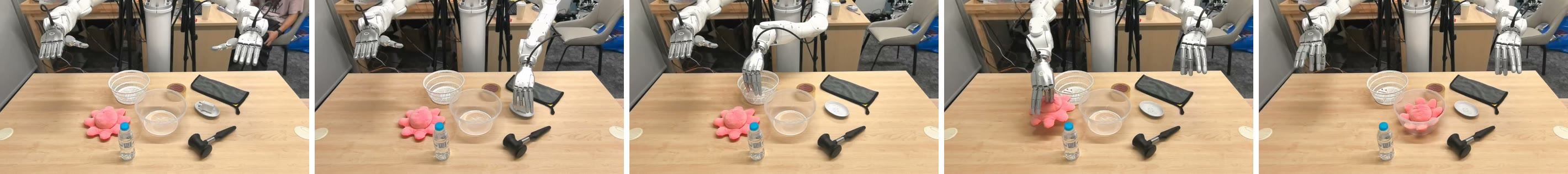}} \\
    Start & Charger plug grasped & Plug into basket & Plush toy grasped & Plush toy in bowl \\
  \end{tabularx}}
  \vspace{-4pt}
  \caption{\textbf{TianJi Marvin humanoid arms.} (a) One hand lifts the lid and holds it while the other puts the card box into the container; the hand holding the lid then closes it. (b) One hand puts the charger plug into the basket, the other the plush toy into the bowl. Each step is checked before it runs, with the same law library as the other robots.}
  \label{fig:marvin}
\end{figure}

\paragraph{Marvin humanoid arms.}
The same law library also drives a TianJi Marvin humanoid with two arms and multi-finger hands (\cref{fig:marvin}). In the first task both hands work together. Putting a card box into a closed container is written as three goals over existing laws: one hand lifts the lid and holds it clear of the opening, the other puts the box inside while the lid stays held, and the hand holding the lid puts it back. In the second task each hand places one object, a charger plug into a basket and a plush toy into a bowl. Each step is checked before it runs; the robot needs only its own profile and adapter. We show these runs qualitatively, without success rates.

\subsection{Capability on LIBERO-PRO}
\label{sec:exp:libero}

LIBERO-PRO~\cite{zhou2025liberopro} perturbs the LIBERO suites so that memorized configurations fail. We use four swap suites (Object, Spatial, Goal, LIBERO-10), 10 tasks each at seeds 0--4, PyRUA-Lean's 200 instances~\cite{si2026pyrua}, scored by the benchmark. All methods use Claude Opus 5.5; alone, it maps images and end-effector state to end-effector actions; PyRUA-Lean writes Python over primitives, including a $\pi_{0.5}$ policy fine-tuned on LIBERO; \textbf{LEAP} uses no learned policy, open loop (one execution) or with one replan.

\begin{table}[t]
  \centering
  \footnotesize
  \setlength{\tabcolsep}{2pt}
  \begin{tabularx}{\textwidth}{@{}l<{\hspace{1.2em}}l<{\hspace{1.2em}}*{5}{>{\centering\arraybackslash}X}@{}}
    \toprule
    Method & Acts via & Object & Spatial & Goal & LIBERO-10 & All \\
    \midrule
    Opus 5.5 & EEF & 56.5 & 90.0 & 45.5 & 0.0 & 48.0 \\
    PyRUA-Lean~\cite{si2026pyrua} & Code\,+\,$\pi_{0.5}$ & \best{100.0} & 80.0 & \best{100.0} & 68.0 & 87.0 \\
    \midrule
    \grouphead{7}{Lean-as-Policy}
    LEAP, open loop & Lean & 60.0 & 82.0 & 70.0 & 34.0 & 61.5 \\
    \ours LEAP, one replan & Lean & \best{100.0} & \best{96.0} & \best{100.0} & \best{84.0} & \best{95.0} \\
    \bottomrule
  \end{tabularx}
  \caption{\looseness=-1\relax \textbf{LIBERO-PRO.} Success rate (\%) on 50 instances per suite. All methods use Claude Opus 5.5; Acts via: end-effector targets (EEF), code with fine-tuned $\pi_{0.5}$, or Lean.}
    \label{tab:libero}
\end{table}

Acting directly, the model succeeds on 48.0\% of the instances and on none of the LIBERO-10 tasks (\cref{tab:libero}). Writing the feasible set instead, LEAP reaches 61.5\% open loop and 95.0\% with one replan, above PyRUA-Lean (87.0\%), which relies on a learned policy. The gains are largest on LIBERO-10 (0.0 to 84.0) and Goal (45.5 to 100.0), whose tasks have several steps or a changed target, since LEAP executes an action only if it satisfies every goal row; one replan corrects 67 of the 77 open-loop failures. The rows compare complete systems that share the model with consistent perception, primitives, and budget.

\begin{table*}[t]
  \centering
  \footnotesize
  \setlength{\tabcolsep}{4.5pt}
  \begin{adjustbox}{max width=\textwidth}\begin{tabular}{@{}llccccccc@{}}
    \toprule
    & & & \multicolumn{1}{c}{Overall} & \multicolumn{5}{c}{SR per dimension (\# tasks)} \\
    \cmidrule(lr){4-4} \cmidrule(l){5-9}
    Method & Type & Zero-shot & SR & Gen. (12) & Prec. (8) & Long-Hor. (8) & Memory (6) & Open (8) \\
    \midrule
    \grouphead{9}{Learned policies}
    $\pi_{0.5}$~\cite{intelligence2025pi05} & VLA & \xmark & 6.93 & 8.17 & 5.50 & 14.67 & 4.67 & 1.67 \\
    VPP2~\cite{guo2026vpp2} & WAM & \xmark & 32.26 & 29.50 & 35.17 & 37.00 & 55.56 & 4.08 \\
    WALL-OSS-1.0-lite & VLA & \xmark & 34.97 & 37.94 & 29.08 & 32.25 & 56.00 & 19.58 \\
    \midrule
    \grouphead{9}{Agent harnesses}
    GPT-6 Astra$^b$~\cite{zhang2026astra} & Agent + EEF & \cmark & 22.48 & 30.50 & 4.00 & 8.25 & 38.67 & 31.00 \\
    PhysicalRSI & Agent + VLA & \xmark & 31.38 & 15.62 & 32.50 & 37.33 & 46.56 & 24.92 \\
    HexaShell-Preview & Agent + Code & \xmark & 46.85 & 34.50 & 32.00 & 31.50 & 81.00 & 55.25 \\
    PyRUA-Lean$^a$~\cite{si2026pyrua} & Agent + Code & \cmark & 50.35 & 43.17 & 38.25 & 43.75 & 68.33 & 58.25 \\
    GO-3 Nano Preview & Agent + WAM & -- & 56.42 & 33.66 & 76.00 & 53.50 & 63.67 & 55.25 \\
    Q0-Preview & Agent + Planner & \cmark & 68.87 & 58.67 & 58.50 & 58.50 & \best{93.67} & \best{75.00} \\
    COAP$^c$~\cite{hu2026etm} & Agent + Code & \xmark & 70.24 & 65.7 & 75.1 & 56.4 & 89.9 & 64.1 \\
    \midrule
    \grouphead{9}{Lean-as-Policy}
    \ours LEAP (ours) & Agent + Lean & \cmark &	\best{73.8} & \best{72.5} & \best{78.8} & \best{61.3} & 91.7 & 70.0 \\
    \bottomrule
  \end{tabular}\end{adjustbox}
  \caption{\textbf{Results on RoboDojo-Sim.} SR: success rate (\%). Zero-shot: \cmark\ built without RoboDojo demonstrations or episodes, \xmark\ trained or developed on them, -- not reported. LEAP: one seed, 10 episodes per task. Baselines: public leaderboard~\cite{robodojo2026board} (Nov.\ 3, 2026), not rerun. $^a$Board evaluation; its paper reports 58.2 / 51.6~\cite{si2026pyrua}. $^b$One end-effector target per model call~\cite{zhang2026astra}. $^c$Author-reported result, averaged over three seeds.}
    \label{tab:main}
\end{table*}

\subsection{From robot data to Lean targets}
\label{sec:exp:data}

Learning the formulation step (\cref{sec:method}) needs pairs of observations and Lean specifications, which recorded robot data lack. We therefore convert each decision of a recorded episode into a Lean target, by rule and without a learned model: the dataset manifest and embodiment table fix each channel's meaning, unit, and frame; the fields observed by then become Lean bindings; the instruction and the annotated sub-tasks not yet finished become typed goals; and Lean elaborates the result against the \texttt{Feasible} library (\cref{fig:data}a,b). Recorded actions never enter the target; they are only checked against it.

\paragraph{Checking recorded episodes.}
\looseness=-1\relax If a target is right, the demonstrations of its task should satisfy it. We test this on six AgiBot tasks, against other episodes with the same sub-tasks. For an episode, the facts are its first observations and the action is its recorded motion: the robot state every 10 frames, the gripper closing and opening at the annotated sub-task boundaries, and a held object moving with the wrist. The episode passes if this action satisfies every row. Three corrections apply to all tasks; two negative controls fail; hand-over task 508 is excluded.

Two values of a target appear in no record, so the target leaves them as Lean placeholders to fit (\cref{fig:data}b,c): the largest step of an arm, \texttt{?s}, and the half-extent \texttt{?h} of a container that receives several objects. \texttt{?s} is the largest step seen in the other tasks, 10\,cm per third of a second. \texttt{?h} is fitted on $N$ calibration episodes of the task, the target's own and $N{-}1$ never used for testing: we centre the container at each episode's first release, measure how far each release lands from it, and take per axis the larger of the farthest release and the mean plus 2.58 standard deviations (a 99\% bound). Checked episodes are centred the same way. Grasp and release times may also shift by up to 0.5\,s, a tolerance chosen after a first run. Tasks 532, 543, and 597, which need no container, then accept 100\%, 99.5\%, and 100\% of unseen episodes, and tasks 709 and 503 accept 87.3\% and 66.7\% with $N = 3$ and 97.9\% and 98.0\% with $N = 100$ (\cref{fig:data}c). Of 400 task-709 episodes with other objects, only 88.2\% pass, because the fitted container does not transfer.

\paragraph{What the rejections are.} Of 58 rejections, 15 are corrupt recordings (\cref{fig:data}d), 40 values out of range and 3 annotation offsets; the 28 with video all succeed: a rejection marks the record, not the demonstration.

\subsection{Standard benchmarks}
\label{sec:exp:bench}

\paragraph{RoboDojo.}
\looseness=-1\relax RoboDojo~\cite{chen2026robodojo} has 42 bimanual tasks for a dual-arm ARX X5 station in five dimensions (\cref{tab:main}), each with 50 public layouts per seed, half of the Generalization layouts randomized. We run one seed with 10 episodes per task (420 episodes) with the official client, judges, and scoring; an abstention, where no plan is admitted, counts as a failure. LEAP receives what every entry receives: three RGB images, the joint and end-effector state of both arms, the instruction, the camera calibration, and the arm mounts; it uses no simulator depth or state and no success signal. LEAP acts repeatedly, one arm per family, executing only kernel-certified plans. LEAP reaches 73.8\% SR, 4.9 points above the best zero-shot entry, Q0-Preview, and 3.6 above COAP, the highest-scoring baseline. It leads on Generalization, Precision, and Long-Horizon, and trails Q0-Preview on Memory (91.7 vs.\ 93.67) and Open (70.0 vs.\ 75.00).

\begin{table}[t]
  \centering
  \footnotesize
  \setlength{\tabcolsep}{3.4pt}
  \begin{tabularx}{\textwidth}{@{}l<{\hspace{1.2em}}l<{\hspace{1.2em}}*{4}{>{\centering\arraybackslash}X}@{}}
    \toprule
    Method & Type & Atomic & Seen & Unseen & Overall \\
    \midrule
    $\pi_{0.5}$~\cite{intelligence2025pi05} & VLA & 39.6 & 7.1 & 1.2 & 16.9 \\
    RLDX-1 & VLA & 67.6 & 27.9 & 8.5 & 36.0 \\
    Xiaomi-Robotics-1 & VLA & 80.2 & 57.1 & 32.1 & 57.4 \\
    Awomo-0.5I & WAM & 86.5 & 58.8 & 45.2 & 64.4 \\
    HexaShell & Agent & 91.1 & 85.0 & 80.0 & 85.6 \\
    \midrule
    \ours LEAP (ours) & Agent + Lean & \best{95.1} & \best{90.5} & \best{84.0} & \best{90.1} \\
    \bottomrule
  \end{tabularx}
  \caption{\textbf{RoboCasa365 target tasks.} Success rate (\%); Seen and Unseen are composite tasks; 50 rollouts per task. Baselines are from the public leaderboard~\cite{robocasa2026board} (Nov.\ 16, 2026), not rerun.}
    \label{tab:robocasa}
\end{table}

\paragraph{RoboCasa365.}
\looseness=-1\relax RoboCasa365~\cite{nasiriany2026robocasa365} places a Panda arm on a wheeled base with a torso lift in simulated kitchens. Of its 50 target tasks, 18 atomic and 16 composite ones are seen in pretraining and 16 composite are unseen; following the official protocol, we run 50 rollouts per task in the pretraining kitchens, scored by the benchmark (\cref{tab:robocasa}). LEAP receives three RGB views and the base, end-effector, and gripper state, without task progress or memory across episodes, and replans after each execution. LEAP reaches 90.1\%, 4.5 points above HexaShell, and it leads on every split.

\section{Discussion}
\label{sec:discussion}

\paragraph{How does LEAP relate to Code as Policies?}
Both have a language model write code, but the code plays a different role. In Code as Policies~\cite{liang2023cap} and its successors~\cite{singh2023progprompt,fu2026capx,si2026pyrua,hu2026etm}, the program is the policy: it calls perception and control functions one after another, so it encodes a single procedure, and whether that procedure was right is seen only after it runs. In LEAP the model writes a specification instead of a procedure: a Lean family that states which actions are valid without choosing one. A solver then picks an action from the family, and the action is checked against the same family before the robot moves. The two approaches can work together. A code-as-policy program can be one more source of candidates for LEAP's sampler, and a LEAP family can judge the program's action before it runs, as it judges the recorded trajectories of \cref{sec:exp:robot}. The difference matters most when the scene is uncertain or the task is new. A procedure covers only the cases its author foresaw, while a specification only has to say what success and safety require, and the search then finds an action that fits the scene actually in front of the robot.

\paragraph{How does LEAP relate to inverse kinematics?}
Inverse kinematics answers one question: which joint angles put the hand at a given pose. LEAP asks a larger one: which poses, paths, and contacts are valid for the task, and whether a given set of joint angles is one of them. Inverse kinematics is the special case with a single goal, the hand at the pose, and the joint limits as the only other constraints. LEAP therefore uses inverse kinematics as a tool, not as the policy. In the sampler, damped least-squares inverse kinematics proposes joint angles for each waypoint, and Lean then checks the rounded angles with exact forward kinematics together with every other law of the task. Three details keep the two apart. The robot executes the angles that were checked, not a fresh solution that could land on another wrist branch. A target outside the arm's reach is rejected by a proved bound before inverse kinematics runs. And when inverse kinematics fails to converge, the result is reported as ``no candidate found'', never as proof that the target cannot be reached.

\paragraph{Why Lean?}
We need a language in which a model can write a specification, a machine can check candidates against it quickly, and a proof can say what a passing check means. Few languages do all three. Python and similar languages run checks quickly, but cannot state or prove what a check guarantees. Classical proof assistants prove such statements, but are rarely used to run a fast checker inside a robot's control loop. Solver input languages, such as SMT, PDDL, or temporal logics, come with mature solvers, but restrict what a constraint can say and leave the soundness of each encoding to the user. Lean~4~\cite{demoura2021lean4} is both a programming language and a proof assistant, so one definition serves as the specification, compiles into the checker, and carries the theorem about it. Its types catch errors such as rows that mix units before anything runs, and its integer and rational arithmetic makes every check exact. Lean also suits the author we have in mind: language models already write Lean well enough to produce machine-checked proofs~\cite{yang2023leandojo,hubert2025alphaproof,ren2025deepseekproverv2}. The costs are real. The laws must be compiled before they run, so we compile them once into a long-lived checker that answers each check quickly; adding or changing a law means rebuilding it. A task can only be stated with laws the library contains, and the proofs cover the stated models and checked points, not the physical world.

\vspace{-4pt}
\section{Conclusion}
\label{sec:conclusion}

We presented LEAP, which writes the set of valid actions as a Lean family over physical laws, lets an objective pick one member, and verifies it against the same specification before the robot moves, stating what the verification covers. LEAP raises the same model from 48.0\% of the LIBERO-PRO instances to 95.0\%, reaches 73.8\% SR on RoboDojo and 90.1\% on RoboCasa365, and on three real tasks completes 13 of 15 rollouts and judges all 30 recorded trajectories correctly. Each executed action comes with the rows and beliefs that admitted it, so a failure can be traced to its cause. Lean targets converted from robot data (\cref{sec:exp:data}) open a path to learning the formulation itself. Because the feasible set can judge actions from any source, it can also filter demonstrations, check learned policies, or train them.

{
    \footnotesize
    \bibliographystyle{ieeenat_fullname}
    \bibliography{main}
}

\end{document}